\pdfoutput=1
\documentclass[letterpaper, 10pt, journal, twoside]{IEEEtran}
\IEEEoverridecommandlockouts

\usepackage[dvipsnames]{xcolor}

\usepackage{amsmath,amsfonts}
\usepackage{algorithmic}
\usepackage{algorithm}
\usepackage{array}
\usepackage{textcomp}
\usepackage{stfloats}
\usepackage{url}
\usepackage{verbatim}
\usepackage{cite}
\usepackage{subcaption}
\usepackage{mathtools} 
\usepackage{overpic}
\usepackage{graphicx}
\usepackage{multirow}
\usepackage{booktabs}
\usepackage{tikz}
\usepackage{pgfplots}
\usetikzlibrary{spy,calc}
\usepackage{hyperref}
\usepackage{pdfpages}
\usepackage{textcomp}
\usepackage{placeins}

\pgfplotsset{compat=1.18} 
\usepackage[normalem]{ulem}

\title{Refining Ground Truth Poses in Autonomous Driving Datasets via Neural Rendering}

\author{Quentin Herau$^{1,2}$, Nathan Piasco$^{1}$, Moussab Bennehar$^{1}$, Luis Roldão$^{1}$, Dzmitry Tsishkou$^{1}$, Bingbing Liu$^{1}$, Cyrille Migniot$^{3}$, Pascal Vasseur$^{4}$ and Cédric Demonceaux$^{2}$%
\thanks{Manuscript received: April 13, 2026; Revised: July 15, 2026; Accepted: August 17, 2026.}%
\thanks{This paper was recommended for publication by Editor Joerg Stueckler upon evaluation of the Associate Editor and Reviewers' comments.}%
\thanks{$^{1}$Quentin Herau, Nathan Piasco, Moussab Bennehar, Luis Roldão, Dzmitry Tsishkou and Bingbing Liu are with Noah's Ark Lab, Huawei {\tt\footnotesize \{quentin.herau, nathan.piasco, moussab.bennehar, luis.roldao, dzmitry.tsishkou, liu.bingbing\}@huawei.com}}%
\thanks{$^{2}$Quentin Herau and Cédric Demonceaux are with Université Bourgogne Europe, CNRS, ICB UMR 6303, Dijon, France {\tt\footnotesize \{quentin.herau@etu., cedric.demonceaux@\}ube.fr}}%
\thanks{$^{3}$Cyrille Migniot is with Université Bourgogne Europe, ImViA UR 7535, Dijon, France {\tt\footnotesize cyrille.migniot@ube.fr}}%
\thanks{$^{4}$Pascal Vasseur is with MIS UR 4290, Université de Picardie Jules Verne, Amiens, France {\tt\footnotesize pascal.vasseur@u-picardie.fr}}%
\thanks{Digital Object Identifier (DOI): see top of this page.}}

\begin{document}
\bstctlcite{IEEEexample:BSTcontrol}

\markboth{IEEE Robotics and Automation Letters. Preprint Version. Accepted September, 2026}%
{Herau \MakeLowercase{\textit{et al.}}: Refining Ground Truth Poses in Autonomous Driving Datasets via Neural Rendering}

\maketitle

\renewcommand{\thefootnote}{\fnsymbol{footnote}}

\begin{abstract}
Public autonomous driving datasets underpin the training and benchmarking of perception, mapping, and localization algorithms, yet residual inaccuracies in sensor calibration and ego-poses can silently degrade both model performance and evaluation reliability.
We introduce MOISST++, a Neural Radiance Field (NeRF)-based pipeline that jointly refines extrinsic sensor calibration and continuous-time ego-trajectories at dataset scale.
The method optimizes shared rig parameters across multiple subsequences and corrects per-subsequence trajectories via a learned continuous-time correction, going beyond prior work that targets individual scenes.
We validate pose improvements without ground truth through a complementary evaluation suite combining Structure from Motion (SfM) triangulation, novel view synthesis, and multi-modal geometric consistency metrics, verify their coherence via cross-metric agreement, and confirm their sensitivity through a controlled-perturbation study with known injected errors.
Applied to four major datasets (KITTI-360, nuScenes, PandaSet, and Waymo), MOISST++ yields statistically significant improvements on most metrics on nuScenes, PandaSet and Waymo, and marginal, within-noise changes on the already well-calibrated KITTI-360.
We publicly release the optimized poses and calibration parameters, together with our evaluation code, to support more reliable research and benchmarking.
\end{abstract}

\section{Introduction}
\IEEEPARstart{P}{ublic} driving datasets~\cite{liao2022kitti,xiao2021pandaset,caesar2020nuscenes,sun2020scalability} are widely used to train and benchmark autonomous driving algorithms, such as perception, mapping and localization.
Their utility relies crucially on the accuracy of the provided vehicle ego-trajectory and the extrinsic sensor calibration.

As the field matures, incremental gains on standard benchmarks (e.g., on the KITTI odometry leaderboard\footnotemark) may become marginal, approaching the accuracy limits imposed by the underlying annotations.
This motivates reassessing the pose and calibration quality of public datasets and providing refined annotations to enable more reliable training and evaluation.

Such inaccuracies in ego poses and sensor calibration can have cascading impact on downstream tasks.
\footnotetext{\url{https://www.cvlibs.net/datasets/kitti/eval_odometry.php}}%
Xia et al.~\cite{xia2025robust} show that a deviation of only 5\,mrad in angular calibration results in a 2.25\,m displacement at 450\,m range, enough to place a vehicle in the wrong lane.
Methods like Neural Radiance Fields (NeRF)~\cite{mildenhall2021nerf}, which rely heavily on accurate pose data for high-fidelity 3D scene representation and novel view synthesis (NVS), can suffer from degraded performance when poses are erroneous.
These inaccuracies not only degrade the performance of downstream methods but also compromise fair and accurate evaluation and benchmarking of different methodologies.

\begin{figure}[!t]
\centering
    Original poses\\[1pt]
    \includegraphics[width=\linewidth]{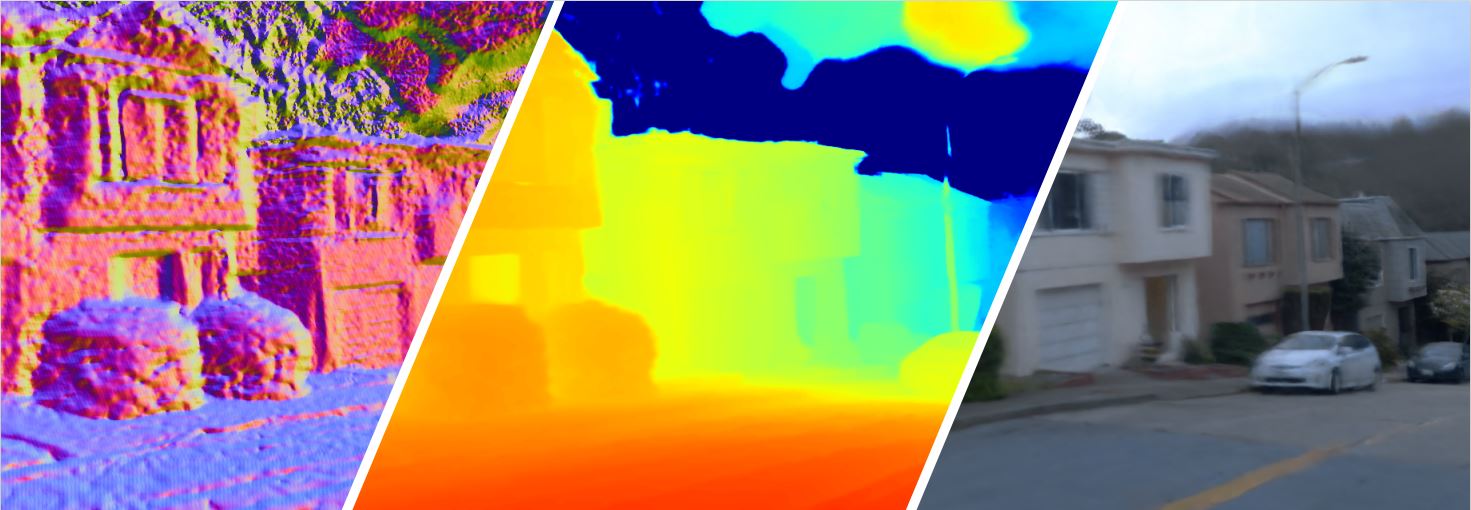}\\[4pt]
    Optimized poses~(Ours)\\[1pt]
    \includegraphics[width=\linewidth]{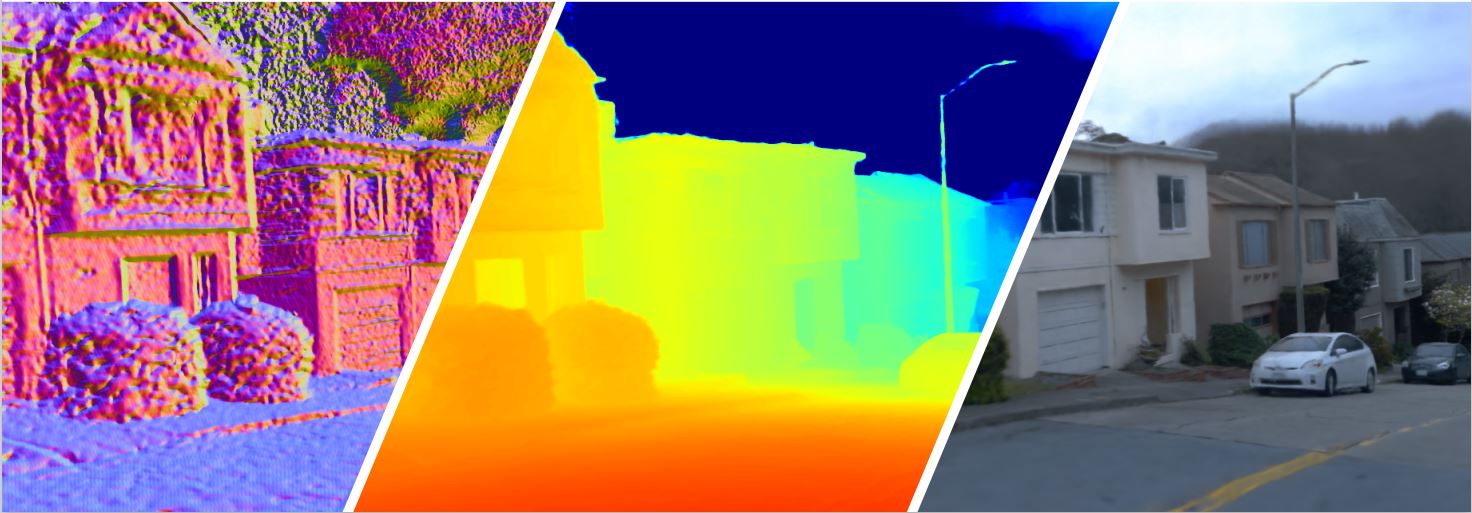}
\caption{\textbf{Impact of our optimized poses for downstream applications.} Rendering comparison between original poses (top) and optimized poses (bottom) on the Waymo Open Dataset~\cite{sun2020scalability} (from left to right: normal map, depth map and RGB rendering). Refined poses yield sharper geometry and fewer pose-induced artifacts; the corresponding quantitative gains across all four datasets are reported in Table~\ref{tab:metrics}.}
\label{fig:datasets}
\end{figure}

Recently, neural rendering approaches have been leveraged to improve geometric and photometric consistency between sensor data (e.g., images and LiDAR scans) and learned 3D models~\cite{wang2021nerf,lin2021barf,herau2023moisst,herau2024soac,yang2025unical}.
Among these, MOISST~\cite{herau2023moisst} and SOAC~\cite{herau2024soac} backpropagate rendering losses to the sensor poses under rigid-rig constraints, enabling robust extrinsic calibration in uncontrolled environments.
However, these methods were designed for sensor calibration on individual scenes and do not directly address the problem of dataset-scale trajectory refinement.

Building on these foundations, we propose a dataset-oriented pipeline that (i)~\emph{estimates a shared extrinsic correction of the sensor rig} across multiple subsequences when applicable, (ii)~\emph{explicitly optimizes a continuous-time ego-trajectory correction} for each subsequence, and (iii)~\emph{provides a multi-metric evaluation framework} to assess pose refinement without requiring ground truth.
In summary, we propose the following contributions in this paper.
\begin{itemize}
    \item We introduce \textbf{MOISST++}, a dataset-refinement adaptation of MOISST that adds \textbf{explicit ego-trajectory optimization} via a continuous-time correction in addition to extrinsic calibration, and apply it at scale across four major autonomous driving datasets.
    \item We introduce a \textbf{complementary evaluation suite} combining triangulation, novel view synthesis, and multi-modal geometric consistency to assess pose refinement without requiring ground truth, and validate its sensitivity through a controlled-perturbation study with known injected errors.
    \item We \textbf{publicly release the optimized poses and our evaluation code} for KITTI-360, nuScenes, PandaSet, and Waymo, providing a ready-to-use resource that lets the community assess pose quality on other datasets.
\end{itemize}

These contributions aim to facilitate more accurate and meaningful research, minimizing errors in model training and evaluation that stem from dataset inaccuracies. By refining both the poses and the evaluation methodology, we obtain improvements on all four datasets, with the largest and statistically significant gains on nuScenes and PandaSet where original calibration and pose inaccuracies are most pronounced, and only marginal changes on the already well-calibrated KITTI-360.

\section{Related Work}
\subsection{Pose and sensor calibration accuracy in driving datasets}
Autonomous driving datasets typically provide sensor extrinsics and vehicle ego-poses obtained via sophisticated proprietary localization pipelines combining multiple sensors, such as GNSS/IMU, LiDAR, and cameras.
Although fairly accurate, residual inaccuracies in poses and calibration remain, and can have a measurable impact on downstream tasks such as 3D reconstruction, sensor fusion, and evaluation.
Rathinam et al.~\cite{rathinam2021recalibrating} demonstrated that recalibrating the KITTI camera setup yields significant improvements in visual odometry accuracy across multiple algorithms, confirming that even well-established datasets can benefit from calibration refinement.
More broadly, sensor misalignment caused by vibrations, thermal effects, or manufacturing tolerances has been shown to severely degrade multi-modal fusion performance, especially at long range~\cite{xia2025robust}.
This motivates post-hoc refinement of provided annotations using self-consistency across multi-view and multi-modal observations.

\subsection{Neural rendering for pose refinement and calibration}
Neural Radiance Fields (NeRF)~\cite{mildenhall2021nerf} leverage differentiable volume rendering to reconstruct 3D scenes and synthesize novel views, while also enabling camera pose refinement through photometric consistency.
BARF~\cite{lin2021barf} and~NeRF$^{--}$~\cite{wang2021nerf} are representative examples of this joint scene reconstruction and pose optimization paradigm.
In the multi-sensor setting, MOISST~\cite{herau2023moisst} introduced rigid rig constraints between sensors to jointly optimize extrinsic calibration and scene representation using both cameras and LiDAR.
SOAC~\cite{herau2024soac} further improved convergence in high-noise regimes by partitioning the scene into sensor-specific representations, reducing interference between modalities.
More recently, UniCal~\cite{yang2025unical} proposed a unified calibration framework using a differentiable surface alignment loss, while UC-NeRF~\cite{cheng2024ucnerf} addressed under-calibrated multi-camera setups with spatiotemporally constrained pose refinement.
In the Gaussian Splatting domain, 3DGS-Calib~\cite{herau20243dgs} achieved comparable calibration accuracy with significantly reduced training time by replacing volume rendering with splatting, and TLC-Calib~\cite{jung2026tlccalib} performed targetless LiDAR-camera calibration by anchoring reliable LiDAR points as Gaussians while refining only the camera extrinsics. As both keep the LiDAR poses fixed, they do not address the dataset-scale trajectory refinement central to our setting.

In this work, we extend MOISST into MOISST++, adding explicit continuous-time ego-trajectory correction and scaling the optimization across multiple subsequences with shared extrinsic parameters.
Unlike prior calibration work, which targets individual scenes, our pipeline is designed for systematic, dataset-wide pose improvement.
We apply the same extension to SOAC, obtaining SOAC++, which we adopt as our neural-rendering competitor: like MOISST++ it leverages both cameras and LiDAR and jointly refines calibration and trajectory. MOISST's simpler formulation is particularly well suited when the original poses and calibration are already reasonably accurate, as for established driving datasets.

\subsection{Pose evaluation without ground truth}
In the absence of ground truth poses, proxy metrics are commonly used to assess pose quality. Multi-view triangulation with SfM toolkits (e.g., COLMAP~\cite{colmap}) provides reprojection error and track statistics that correlate with pose quality. Novel view synthesis (NVS) quality metrics (PSNR, SSIM~\cite{ssim}, LPIPS) are widely used to assess whether optimized poses lead to more consistent renderings. Finally, geometric consistency between reconstructed surfaces and independent depth measurements (e.g., LiDAR) provides an additional modality-agnostic validation signal~\cite{djeghim2025viineus}. We combine these complementary evaluations to reduce metric bias and to better support conclusions about pose refinement quality.

\section{Method}
The goal of this work is to refine (i) the extrinsic calibration of the multi-modal multi-sensor rig and (ii) the ego-trajectory provided in driving datasets, using neural rendering optimization backends. In practice, \emph{sequential} optimization is used: a dataset-level extrinsic correction is first estimated (when the sensor rig is physically fixed across sequences), then the trajectory is refined per subsequence while keeping the extrinsics fixed. The resulting poses are assessed using complementary geometric and photometric criteria (Sec.~\ref{method/evaluations}).

\subsection{Pipeline overview} \label{method/pipeline}

\begin{figure*}[htbp]
  \centering
  \includegraphics[width=0.92\textwidth]{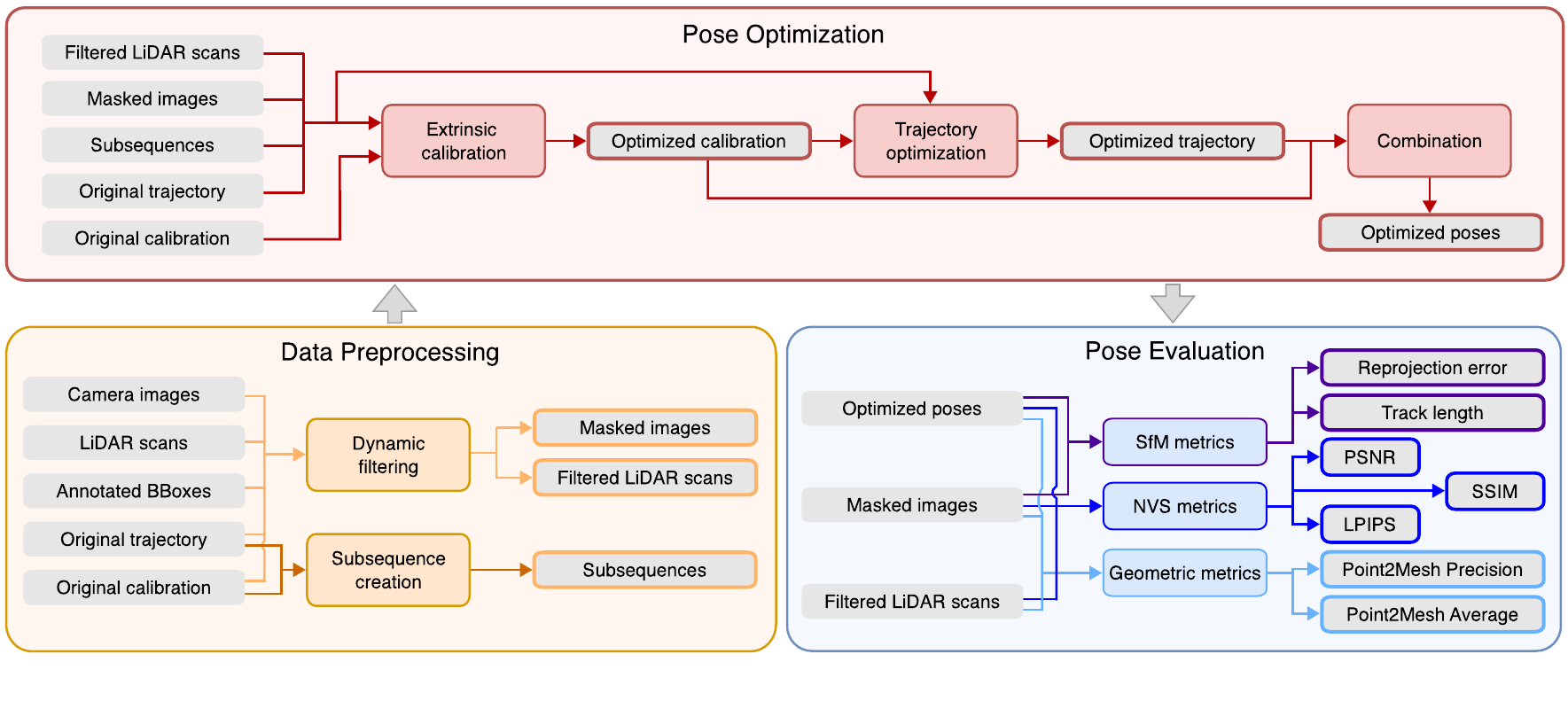}
  \caption{\textbf{Pipeline:}  First, the data is preprocessed to remove dynamic elements, and create subsequences fitting our NeRF models. Then, the extrinsic parameters are optimized, followed by the trajectories. Finally, the new poses are evaluated.} 
  \label{fig:pipeline}
\end{figure*}

The pipeline (Fig.~\ref{fig:pipeline}) consists of all necessary steps to obtain the desired outputs: the optimal calibration parameters and the vehicle's ego-trajectory.
\begin{itemize}
    \item \textbf{Dynamic Filtering}: Since the NeRF models used do not handle dynamic objects, instance-level bounding boxes provided by the datasets are used to filter them out. Dynamic pixels in images are removed by reprojecting tracked boxes, and LiDAR points within these boxes are excluded. Enlarged bounding boxes account for pose noise.

    \item \textbf{Subsequence Creation}: Sequences are split into smaller subsequences of fixed size so that each NeRF model can maintain a consistent level of detail.

    \item \textbf{Extrinsic Calibration Optimization}: Assuming the sensor rig is fixed across sequences, calibration parameters are optimized across multiple subsequences simultaneously to enhance robustness. 

    \item \textbf{Trajectory Optimization}: With calibration fixed, the trajectory is refined independently for each subsequence.

    \item \textbf{Evaluation}: The optimized poses are evaluated via keypoint reprojection, geometric consistency, and NVS (Sec.~\ref{method/evaluations}).
\end{itemize}

\subsection{Optimization process} \label{method/optimization}
\subsubsection{Model}
The pipeline is compatible with any method capable of multi-modal, multi-sensor calibration and trajectory optimization. We use:
\begin{itemize} 
    \item \textbf{MOISST++}: MOISST refines a multi-sensor rig by backpropagating rendering losses to input poses while enforcing rigid constraints between sensors. MOISST++ denotes our dataset-refinement adaptation with synchronized sensors and an explicit \emph{continuous-time} ego-trajectory correction $\mathcal{T}_{\text{cor}}(t)$ optimized in addition to extrinsics.
    \item \textbf{SOAC++}: SOAC partitions the scene into sensor-specific representations to reduce interference between modalities during multi-sensor calibration. SOAC++ denotes its dataset-refinement variant with the same continuous-time trajectory correction $\mathcal{T}_{\text{cor}}(t)$, used as a neural-rendering competitor that also leverages both cameras and LiDAR.
    \item \textbf{Nerfstudio~\cite{tancik2023nerfstudio}}: Nerfacto's built-in pose optimization is used as an optimization baseline that does not rely on rig constraints or LiDAR data. Note that Nerfacto (and Splatfacto) are also used separately as independent NVS evaluators with pose optimization disabled (Sec.~\ref{method/evaluations}).
\end{itemize} 

\subsubsection{Formulation}
We denote $S$ the set of sensors to be calibrated, $\prescript{}{w}{T}^i(t) \!\in\! \mathbf{R}^{4\times 4}$ the pose of sensor~$i \!\in\! S$ at time~$t$ in world coordinates, and $\prescript{}{j}{T}^i \!\in\! \mathbf{R}^{4 \times 4}$ the transformation from sensor~$i$ to sensor~$j$.

The vehicle pose is associated with a reference sensor $r \in S$. A continuous trajectory $\mathcal{T}_r(t)$ is constructed from the discrete poses of $r$ using linear interpolation for translation and SLERP for rotation:
\begin{equation} 
    \prescript{}{w}{T}^r(t) = \mathcal{T}_r(t),
\end{equation} 
and the pose of any sensor $i$ at time $t$ is obtained as: 
\begin{equation} 
    \prescript{}{w}{T}^i(t) = \mathcal{T}_r(t) \; \prescript{}{r}{T}^i.
\end{equation}

To optimize both the extrinsic calibration and trajectory, we define $E_{\text{cor}}^i$ as the extrinsic correction for sensor $i$ and $\mathcal{T}_{\text{cor}}(t)$ as the trajectory correction at time $t$. The optimized pose becomes: 
\begin{equation} 
    \prescript{}{w}{T}^i(t) = \mathcal{T}_{\text{cor}}(t) \; \mathcal{T}_r(t) \; \prescript{}{r}{T}^i \; E_{\text{cor}}^i.
\end{equation}

\subsubsection{Implementation details}
$E_{\text{cor}}$ is parameterized as a 6-dimensional embedding (3 for translation, 3 for rotation via the Rodrigues axis-angle representation), shared across subsequences to encourage convergence on a common calibration.

For $\mathcal{T}_{\text{cor}}$, we adopt the continuous representation of Ma~et~al.~\cite{ma2024continuous}: an MLP learns trajectory corrections as a function of time. To prevent convergence to local minima, weight decay is applied to the NeRF hash grids during initial epochs, following the coarse-to-fine strategy of MOISST~\cite{herau2023moisst}.

During extrinsic optimization, multiple MOISST++ instances run on different subsequences with individual $\mathcal{T}_{\text{cor}}$ but shared $E_{\text{cor}}$. During trajectory optimization, $E_{\text{cor}}$ is frozen and only $\mathcal{T}_{\text{cor}}$ and the scene representation are optimized.

\subsubsection{Optimization objective}
The objective for a subsequence is:
\begin{equation}
\begin{aligned}
    \min_{\theta,\,E_{\text{cor}},\,\mathcal{T}_{\text{cor}}}\;&
    \lambda_{\text{rgb}}\mathcal{L}_{\text{photo}}(\theta,E_{\text{cor}},\mathcal{T}_{\text{cor}})\\
    &+\lambda_{\text{lidar}}\mathcal{L}_{\text{lidar}}(\theta,E_{\text{cor}},\mathcal{T}_{\text{cor}})\\
    &+\lambda_{\text{reg}}\mathcal{L}_{\text{reg}},
\end{aligned}
\end{equation}
where $\theta$ denotes the neural scene parameters, $\mathcal{L}_{\text{photo}}$ enforces multi-view photometric consistency, $\mathcal{L}_{\text{lidar}}$ enforces consistency with LiDAR supervision~\cite{herau2023moisst}, and $\mathcal{L}_{\text{reg}}$ includes distortion and interlevel regularization from Mip-NeRF~360~\cite{barron2022mip}. For calibration, $E_{\text{cor}}$ is shared across subsequences; for trajectory optimization, $E_{\text{cor}}$ is fixed. 

\subsection{Pose accuracy assessment} \label{method/evaluations}

\subsubsection{Triangulation metrics}
Pose quality is assessed using COLMAP~\cite{colmap} keypoint triangulation with SIFT features. Crucially, poses are \emph{not} refined via bundle adjustment, so the metrics directly reflect input pose quality. We report the average reprojection error (in pixels) and track length (number of views per tracked keypoint).

\subsubsection{Novel View Synthesis (NVS)}
We evaluate NVS with two models. \textbf{NeRF-LS} is the NeRF model with LiDAR supervision from~\cite{herau2023moisst} (omitting depth smoothness) and per-camera appearance embeddings inspired by NeRF-in-the-Wild~\cite{martin2021nerf}. \textbf{Nerfacto}~\cite{tancik2023nerfstudio} is evaluated with pose optimization disabled to show that improvements transfer to an independent implementation. Qualitative renderings with \textbf{Splatfacto} illustrate that refined poses also benefit 3D Gaussian Splatting.
PSNR, SSIM and LPIPS are reported for original and optimized poses.

\subsubsection{Geometric consistency}
The Delaunay mesher in COLMAP creates a 3D mesh from the triangulated sparse points, which is compared to the accumulated LiDAR point cloud. Following~\cite{djeghim2025viineus}, the mean point-to-mesh distance (P$\rightarrow$M, in meters) and the fraction of points within 15\,cm (Prec., in $[0,1]$) are reported.

\section{Experimental setup}
\subsection{Datasets}

The pipeline is evaluated on four widely used public autonomous driving datasets:
\subsubsection{KITTI-360}
KITTI-360~\cite{liao2022kitti} extends the original KITTI dataset with 360° panoramic imagery, offering data from 4 cameras, a 32-beam LiDAR, and GPS/IMU sensors in Karlsruhe, Germany.
The provided 6-DoF poses are estimated using a combination of LiDAR-based SLAM and SfM techniques, refined via global optimization and loop closures.

\subsubsection{nuScenes}
nuScenes~\cite{caesar2020nuscenes} provides 1,000 driving scenes (each 20\,s) across Boston and Singapore with 6 cameras, a 32-beam LiDAR, radar, and GPS/IMU.
Vehicle poses are computed by fusing GPS/IMU with an HD map built from LiDAR scans, and are defined in \textbf{SE2} (no Z-axis information).

\subsubsection{PandaSet}
PandaSet~\cite{xiao2021pandaset} covers urban environments with 5 cameras, a 64-beam LiDAR, and radar.
Poses are obtained through localization pipelines integrating LiDAR, cameras, inertial sensors and GNSS, with LiDAR SLAM post-processing for consistency.

\subsubsection{Waymo Open Dataset}
Waymo~\cite{sun2020scalability} contains over 1,000 hours of data from 5 LiDAR sensors and 5 cameras across urban and suburban environments in Phoenix, Arizona.
Waymo uses its in-house localization system, fusing high-resolution LiDAR, IMU, and additional sensors to deliver accurate 6-DoF poses in a global reference frame.

\subsection{Subsequence creation} \label{setup/subseq}
Given the large size of the datasets and computational demands of training NeRF models, results are reported on systematic subsets of each dataset, avoiding hand-picked bias:
\begin{itemize} 
\item \textbf{KITTI-360}: 50\,m subsequences starting from frame 500 of sequences \textbf{0005, 0006, 0007, 0009, 0010} (50 subsequences, 2.5\,km total). 
\item \textbf{nuScenes}: First 30\,m of the first 10 sequences. 
\item \textbf{PandaSet}: First 10 sequences with semantic labels (\textbf{001, 002, 003, 005, 011, 013, 015, 016, 017, 019}).
\item \textbf{Waymo}: First 50\,m from 10 static sequences identified via annotated bounding boxes (\textbf{segment-10061, -10676, -11379, -11724, -12879, -13085, -13142, -13196, -14004, -14869}). 
\end{itemize}

For each dataset, camera and 360° LiDAR poses are optimized. Dynamic masks are generated for KITTI-360, nuScenes, and PandaSet using the provided annotations. For Waymo, different capture vehicles lead to varying extrinsics, so calibration and trajectory are optimized per subsequence individually (see Sec.~\ref{method/pipeline}).

\subsection{Optimization hyperparameters} \label{experiments/optimization_hyperparameters}
The image downscale factors and iteration counts are detailed in Table~\ref{tab:hyperparameters}. 
\begin{table}[h]
    \centering
        \caption{Hyperparameters: images are downscaled during optimization to reduce training time and memory usage.}
    \label{tab:hyperparameters}

    \footnotesize
    \renewcommand{\arraystretch}{0.95}
    \begin{tabular}{lll}
    \toprule
    \textbf{Stage} & \textbf{Parameter} & \textbf{Value} \\
    \midrule
    \multirow{2}{*}{Optimization} & Downscale factor & 2 (KITTI-360), 4 (others) \\
     & Iterations & 15 epochs \\
    \midrule
    \multirow{2}{*}{NeRF-LS (NVS)} & Downscale factor & 1 (KITTI-360), 2 (others) \\
     & Iterations & 10 epochs \\
    \midrule
    \multirow{2}{*}{Nerfstudio (NVS)} & Downscale factor & 1 \\
     & Iterations & 30k steps \\
    \bottomrule
    \end{tabular}

\end{table}

\subsection{Compute and scalability}
All experiments run on a single NVIDIA RTX 4090 GPU. End-to-end optimization of one subsequence takes \textbf{$\sim$30 minutes} (including NeRF training and pose refinement). Runtime scales linearly with the number of subsequences; the calibration stage can be parallelized across subsequences when optimizing shared extrinsics. For the largest evaluation (KITTI-360, 50 subsequences), total wall-clock time is approximately 25 hours when run sequentially. The pipeline requires no manual tuning per dataset beyond the hyperparameters listed in Table~\ref{tab:hyperparameters}, making it practical for large-scale deployment.

\section{Experimental Results}
The poses obtained through the pipeline using MOISST++ are evaluated with the metrics of Sec.~\ref{method/evaluations}.
For \textit{nuScenes}, the provided poses lack Z-axis information. To initialize an approximate Z component, we run KISS-ICP~\cite{vizzo2023kiss}. Table~\ref{tab:icp} shows that most of the gain comes from MOISST++ rather than this initialization.

\begin{table}[!htbp]
    \centering
        \caption{nuScenes metrics: KISS-ICP~\cite{vizzo2023kiss} improves compared to Original, while still providing inferior metrics to MOISST++. The best result in \textbf{bold}, the second best \underline{underlined}.}
    \label{tab:icp}

    \footnotesize
    \renewcommand{\arraystretch}{0.95}
        \begin{tabular}{l l ccc}
        \toprule
        \multicolumn{2}{c}{Metric} & Original & KISS-ICP & MOISST++\\
        \cmidrule(l{1pt}r{1pt}){1-2} \cmidrule(l{1pt}r{1pt}){3-5}
        \multirow{2}{*}{\textbf{Colmap}} 
        & Reproj. Error~$\downarrow$&1.339&\underline{1.306}&\textbf{0.852}\\
        &Track Length~$\uparrow$&5.40&\underline{5.41}&\textbf{7.20}\\
        \midrule
        \multirow{3}{*}{\textbf{NeRF-LS}} 
        &PSNR~$\uparrow$&22.34&\underline{22.76}&\textbf{23.46}\\
        &SSIM~$\uparrow$&0.810&\underline{0.823}&\textbf{0.851}\\
        &LPIPS~$\downarrow$&0.465&\underline{0.418}&\textbf{0.339}\\
        \midrule
        \multirow{3}{*}{\textbf{Nerfacto}} 
        & PSNR~$\uparrow$&20.41&\underline{20.54}&\textbf{21.68}\\
        &SSIM~$\uparrow$&0.696&\underline{0.701}&\textbf{0.778}\\
        &LPIPS~$\downarrow$&0.545&\underline{0.531}&\textbf{0.400}\\
        \midrule
        \multirow{2}{*}{\textbf{Delauney}} 
        & Precision~$\uparrow$&0.422&\underline{0.508}&\textbf{0.673} \\
        &P$\rightarrow$M~[m]~$\downarrow$&0.296&\underline{0.171}&\textbf{0.129}\\
        \bottomrule
       \end{tabular}
    
\end{table}

Quantitative results across datasets and metrics are summarized in Table~\ref{tab:metrics}.
\begin{table*}[t]
    \centering
        \caption{Quantitative results: SOAC++ and the camera-only Nerfstudio baseline are compared against MOISST++. Best result in \textbf{bold}; second best \underline{underlined} when applicable.}
\label{tab:metrics}

    \scriptsize
    \setlength{\tabcolsep}{4pt}
    \renewcommand{\arraystretch}{0.95}

    \begin{subtable}[t]{\textwidth}
        \centering
        \caption{COLMAP triangulation metrics}
        \label{tab:colmap_metrics}

        \begin{tabular}{l @{\hspace{8pt}} cccc @{\hspace{10pt}} cccc}
        \cmidrule(l{1pt}r{1pt}){2-9}
        & \multicolumn{4}{c}{Reprojection Error~$\downarrow$} & \multicolumn{4}{c}{Track Length~$\uparrow$} \\
        \cmidrule(l{2pt}r{2pt}){2-5} \cmidrule(l{2pt}r{2pt}){6-9}
        Dataset & Original & Nerfstudio & SOAC++ & MOISST++ & Original & Nerfstudio & SOAC++ & MOISST++\\
        \midrule
        KITTI-360 & \underline{0.586} & 0.691 & 0.735 & \textbf{0.577} & 6.53 & \underline{6.55} & 6.48 & \textbf{6.68}\\
        nuScenes  & 1.339 & 1.166 & \underline{1.016} & \textbf{0.852} & 5.40 & 6.59 & \underline{6.60} & \textbf{7.20}\\
        PandaSet  & 1.187 & 1.284 & \underline{1.020} & \textbf{0.981} & 6.32 & 6.44 & \underline{6.75} & \textbf{6.86}\\
        Waymo     & \underline{1.137} & 1.466 & 1.344 & \textbf{1.015} & \underline{5.73} & 5.51 & 5.19 & \textbf{6.11}\\
        \bottomrule
        \end{tabular}
        
    \end{subtable}

    \vspace{7pt}

    \begin{subtable}[t]{\textwidth}
        \centering
        \caption{NVS metrics (NeRF-LS)}
        \label{tab:NVS_imgine}

        \begin{tabular}{l @{\hspace{8pt}} ccc @{\hspace{10pt}} ccc @{\hspace{10pt}} ccc}
        \cmidrule(l{1pt}r{1pt}){2-10}
        & \multicolumn{3}{c}{PSNR~$\uparrow$} & \multicolumn{3}{c}{SSIM~$\uparrow$} & \multicolumn{3}{c}{LPIPS~$\downarrow$} \\
        \cmidrule(l{2pt}r{2pt}){2-4} \cmidrule(l{2pt}r{2pt}){5-7} \cmidrule(l{2pt}r{2pt}){8-10}
        Dataset & Original & SOAC++ & MOISST++ & Original & SOAC++ & MOISST++ & Original & SOAC++ & MOISST++\\
        \midrule
        KITTI-360 & \underline{22.42} & 22.28 & \textbf{22.61} & \underline{0.829} & 0.824 & \textbf{0.834} & \underline{0.283} & 0.295 & \textbf{0.272}\\
        nuScenes  & 22.34 & \underline{23.12} & \textbf{23.46} & 0.810 & \underline{0.843} & \textbf{0.851} & 0.465 & \underline{0.370} & \textbf{0.339}\\
        PandaSet  & 22.08 & \underline{22.32} & \textbf{22.36} & 0.788 & \underline{0.800} & \textbf{0.803} & 0.482 & \underline{0.441} & \textbf{0.433}\\
        Waymo     & 24.56 & \underline{24.63} & \textbf{25.03} & 0.848 & \underline{0.849} & \textbf{0.857} & 0.468 & \underline{0.466} & \textbf{0.420}\\
        \bottomrule
        \end{tabular}
        
    \end{subtable}

    \vspace{7pt}

    \begin{subtable}[t]{\textwidth}
        \centering
        \caption{NVS metrics (Nerfacto)}
        \label{tab:NVS_Nerfacto}

        \begin{tabular}{l @{\hspace{6pt}} cccc @{\hspace{8pt}} cccc @{\hspace{8pt}} cccc}
        \cmidrule(l{1pt}r{1pt}){2-13}
        & \multicolumn{4}{c}{PSNR~$\uparrow$} & \multicolumn{4}{c}{SSIM~$\uparrow$} & \multicolumn{4}{c}{LPIPS~$\downarrow$} \\
        \cmidrule(l{2pt}r{2pt}){2-5} \cmidrule(l{2pt}r{2pt}){6-9} \cmidrule(l{2pt}r{2pt}){10-13}
        Dataset & Orig. & Nerf. & SOAC++ & MOISST++ & Orig. & Nerf. & SOAC++ & MOISST++ & Orig. & Nerf. & SOAC++ & MOISST++\\
        \midrule
        KITTI-360 & \underline{21.24} & \underline{21.24} & 21.05 & \textbf{21.37} & \underline{0.713} & 0.710 & 0.700 & \textbf{0.720} & \underline{0.281} & 0.284 & 0.294 & \textbf{0.274}\\
        nuScenes  & 20.41 & \underline{21.61} & 21.41 & \textbf{21.68} & 0.696 & \underline{0.765} & 0.756 & \textbf{0.778} & 0.545 & \underline{0.422} & 0.436 & \textbf{0.400}\\
        PandaSet  & 21.60 & 22.14 & \underline{22.34} & \textbf{22.40} & 0.686 & 0.710 & \underline{0.718} & \textbf{0.723} & 0.517 & 0.484 & \underline{0.472} & \textbf{0.461}\\
        Waymo     & 23.54 & \underline{24.40} & 23.95 & \textbf{24.54} & 0.750 & \underline{0.767} & 0.755 & \textbf{0.772} & 0.493 & \underline{0.465} & 0.489 & \textbf{0.450}\\
        \bottomrule
        \end{tabular}
        
    \end{subtable}

    \vspace{7pt}

    \begin{subtable}[t]{\textwidth}
        \centering
        \caption{NVS metrics (Splatfacto, 3D Gaussian Splatting)}
        \label{tab:NVS_splatfacto}

        \begin{tabular}{l @{\hspace{8pt}} ccc @{\hspace{10pt}} ccc @{\hspace{10pt}} ccc}
        \cmidrule(l{1pt}r{1pt}){2-10}
        & \multicolumn{3}{c}{PSNR~$\uparrow$} & \multicolumn{3}{c}{SSIM~$\uparrow$} & \multicolumn{3}{c}{LPIPS~$\downarrow$} \\
        \cmidrule(l{2pt}r{2pt}){2-4} \cmidrule(l{2pt}r{2pt}){5-7} \cmidrule(l{2pt}r{2pt}){8-10}
        Dataset & Original & SOAC++ & MOISST++ & Original & SOAC++ & MOISST++ & Original & SOAC++ & MOISST++\\
        \midrule
        KITTI-360 & \underline{23.40} & 22.99 & \textbf{23.63} & \underline{0.757} & 0.743 & \textbf{0.764} & \underline{0.223} & 0.235 & \textbf{0.216}\\
        nuScenes  & 22.29 & \underline{24.77} & \textbf{25.66} & 0.762 & \underline{0.825} & \textbf{0.852} & 0.397 & \underline{0.297} & \textbf{0.264}\\
        PandaSet  & 23.22 & \underline{24.11} & \textbf{24.31} & 0.770 & \underline{0.802} & \textbf{0.811} & 0.323 & \underline{0.294} & \textbf{0.286}\\
        Waymo     & \underline{24.06} & 23.69 & \textbf{25.85} & \underline{0.788} & 0.777 & \textbf{0.819} & \underline{0.407} & 0.424 & \textbf{0.346}\\
        \bottomrule
        \end{tabular}
        
    \end{subtable}

    \vspace{7pt}

    \begin{subtable}[t]{\textwidth}
        \centering
        \caption{Delaunay mesh metrics}
        \label{tab:delauney_metrics}

        \begin{tabular}{l @{\hspace{8pt}} ccc @{\hspace{10pt}} ccc}
        \cmidrule(l{1pt}r{1pt}){2-7}
        & \multicolumn{3}{c}{Precision~$\uparrow$} & \multicolumn{3}{c}{P$\rightarrow$M [m]~$\downarrow$} \\
        \cmidrule(l{2pt}r{2pt}){2-4} \cmidrule(l{2pt}r{2pt}){5-7}
        Dataset & Original & SOAC++ & MOISST++ & Original & SOAC++ & MOISST++\\
        \midrule
        KITTI-360 & \underline{0.589} & 0.582 & \textbf{0.600} & \underline{0.254} & 0.261 & \textbf{0.228}\\
        nuScenes  & 0.422 & \underline{0.632} & \textbf{0.673} & 0.296 & \underline{0.136} & \textbf{0.129}\\
        PandaSet  & 0.385 & \underline{0.393} & \textbf{0.400} & 0.378 & \underline{0.357} & \textbf{0.355}\\
        Waymo     & \underline{0.441} & 0.417 & \textbf{0.626} & 0.408 & \underline{0.386} & \textbf{0.340}\\
        \bottomrule
        \end{tabular}
        
    \end{subtable}

\end{table*}

\subsection{Triangulation metrics}
Table~\ref{tab:colmap_metrics} reports the COLMAP reprojection error and track length.
MOISST++ yields the best overall improvement. On the already-accurate KITTI-360 the change is minimal (reprojection $0.586\!\to\!0.577$, track length $6.53\!\to\!6.68$), whereas the other datasets improve substantially: the reprojection error drops from $1.339$ to $0.852$ on nuScenes, $1.187$ to $0.981$ on PandaSet, and $1.137$ to $1.015$ on Waymo, with matching track-length gains (e.g.\ $5.40\!\to\!7.20$ on nuScenes and $5.73\!\to\!6.11$ on Waymo). Both indicate more consistent multi-view geometry, i.e.\ higher pose quality.

\subsection{Novel View Synthesis}
\begin{figure}[!htbp]
    \centering
    \setlength{\tabcolsep}{0.0001\linewidth}

    \begin{subfigure}{.48\linewidth}
        \centering
        \resizebox{\linewidth}{!}{%
        \begin{tikzpicture}[outer sep=0pt,inner sep=0pt,spy using outlines={rectangle, connect spies}]
            \node (img) {\includegraphics[width=\linewidth]{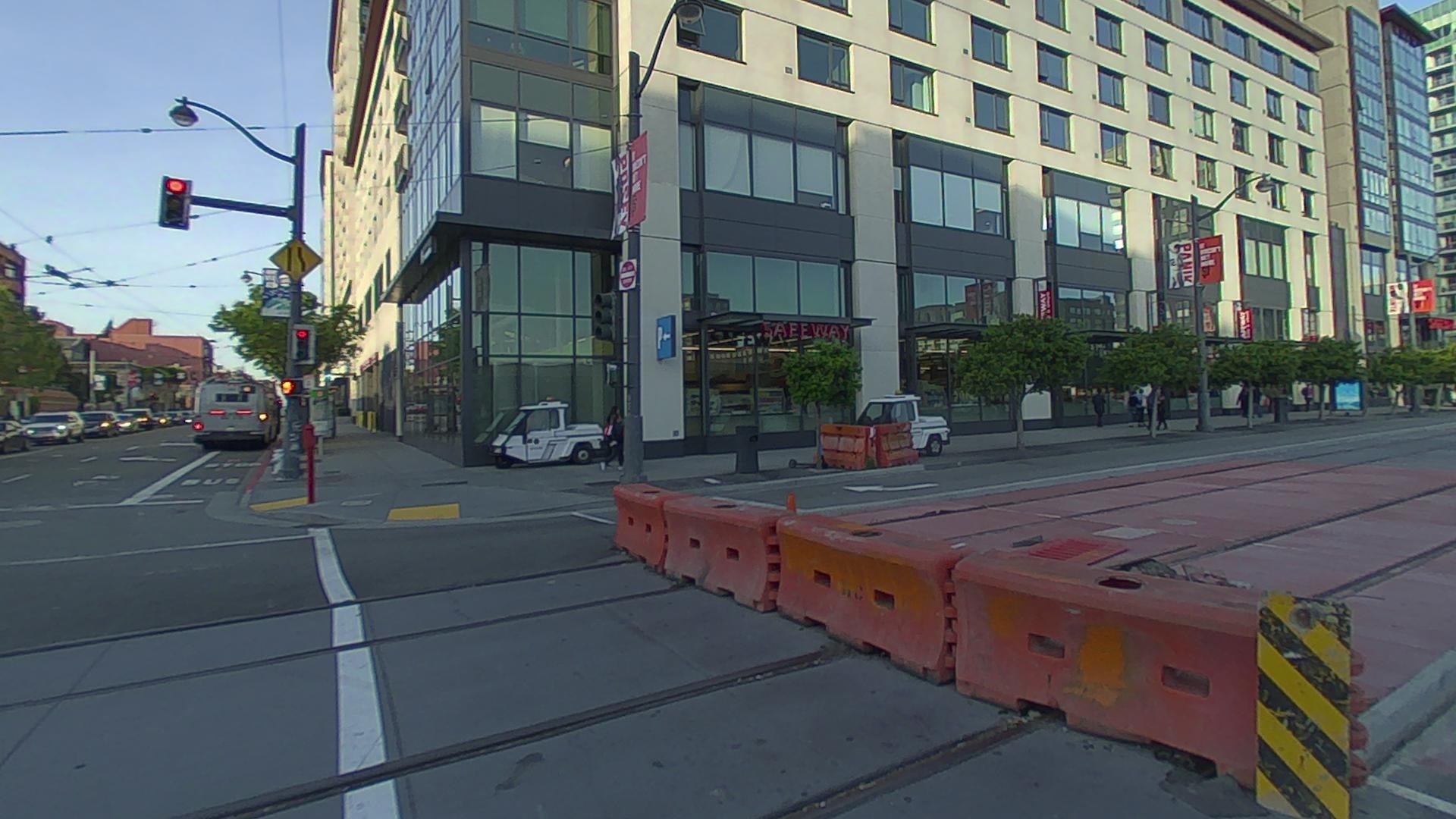}};
            \begin{scope}[shift={(img.south west)},x={(img.south east)},y={(img.north west)}]
                \spy [red,magnification=2.5,size=1.3cm] on (-1.7,0.75) in node at (-1.2,-0.5);
                \spy [orange,magnification=4,size=1cm] on (1.35,0.45) in node at (0.99,-0.5);
            \end{scope}
        \end{tikzpicture}%
        }
        \caption{GT}
        \label{figures/pandaset/qualitative_ns/gt}
    \end{subfigure}

    \vspace{0.3mm}

    \begin{subfigure}{.48\linewidth}
        \centering
        \resizebox{\linewidth}{!}{%
        \begin{tikzpicture}[outer sep=0pt,inner sep=0pt,spy using outlines={rectangle, connect spies}]
            \node (img) {\includegraphics[width=\linewidth]{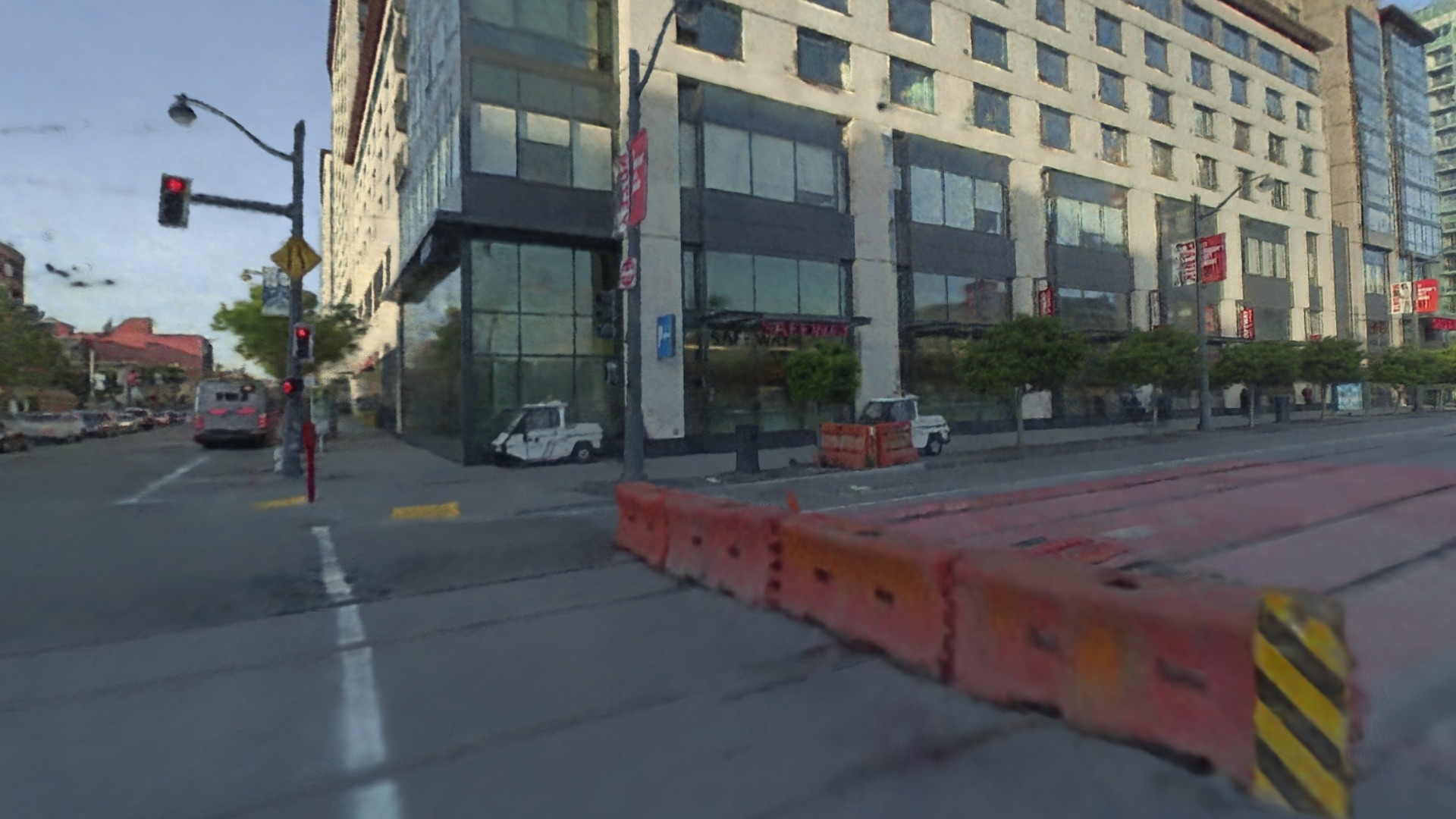}};
            \begin{scope}[shift={(img.south west)},x={(img.south east)},y={(img.north west)}]
                \spy [red,magnification=2.5,size=1.3cm] on (-1.7,0.75) in node at (-1.2,-0.5);
                \spy [orange,magnification=4,size=1cm] on (1.35,0.45) in node at (0.99,-0.5);
            \end{scope}
        \end{tikzpicture}%
        }
        \caption{Original (Nerfacto)}
        \label{figures/pandaset/qualitative_ns/original_nerfacto}
    \end{subfigure}
    \hfill
    \begin{subfigure}{.48\linewidth}
        \centering
        \resizebox{\linewidth}{!}{%
        \begin{tikzpicture}[outer sep=0pt,inner sep=0pt,spy using outlines={rectangle, connect spies}]
            \node (img) {\includegraphics[width=\linewidth]{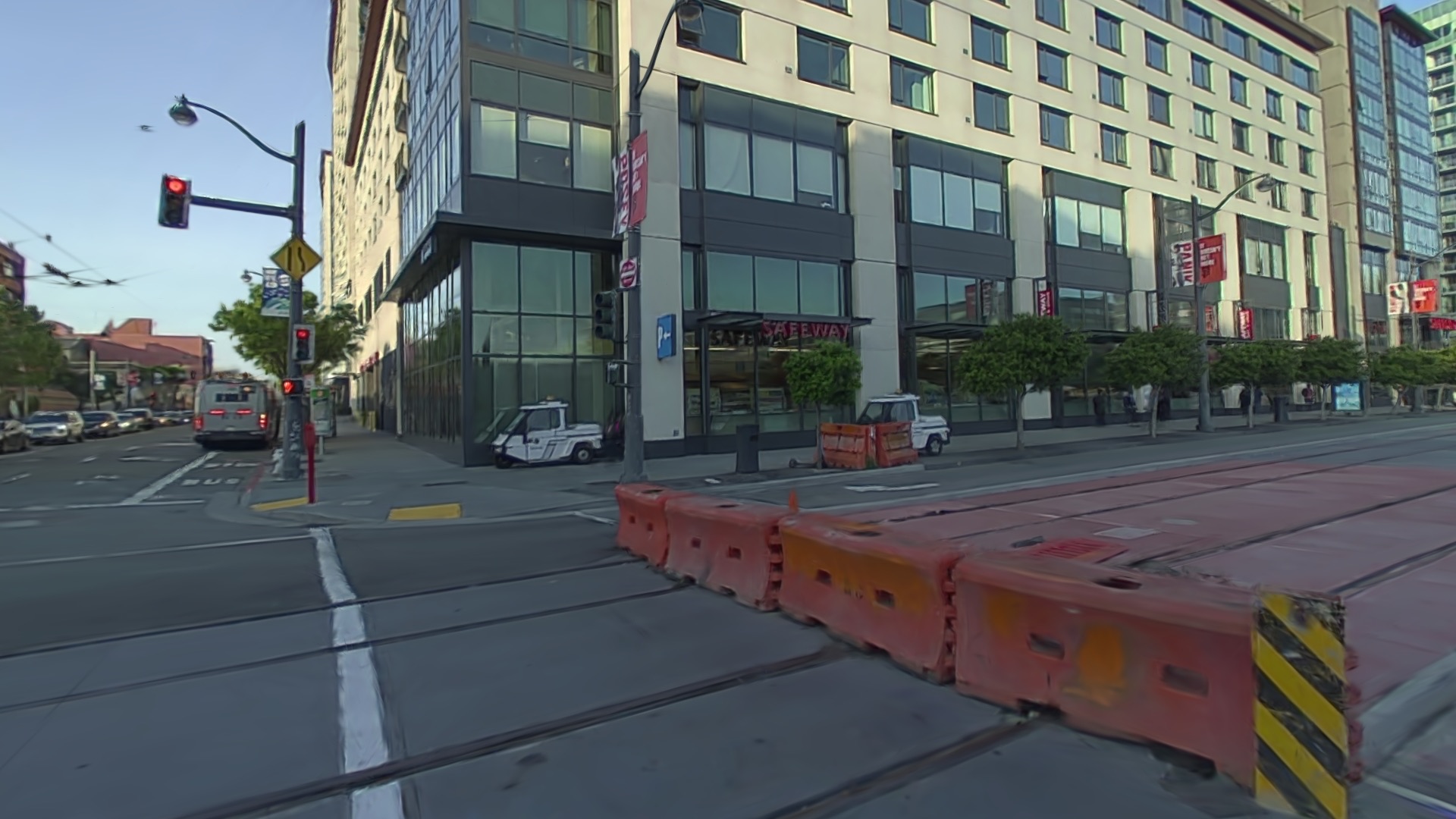}};
            \begin{scope}[shift={(img.south west)},x={(img.south east)},y={(img.north west)}]
                \spy [red,magnification=2.5,size=1.3cm] on (-1.7,0.75) in node at (-1.2,-0.5);
                \spy [orange,magnification=4,size=1cm] on (1.35,0.45) in node at (0.99,-0.5);
            \end{scope}
        \end{tikzpicture}%
        }
        \caption{Original (Splatfacto)}
        \label{figures/pandaset/qualitative_ns/original_splatfacto}
    \end{subfigure}

    \vspace{0.3mm}

    \begin{subfigure}{.48\linewidth}
        \centering
        \resizebox{\linewidth}{!}{%
        \begin{tikzpicture}[outer sep=0pt,inner sep=0pt,spy using outlines={rectangle, connect spies}]
            \node (img) {\includegraphics[width=\linewidth]{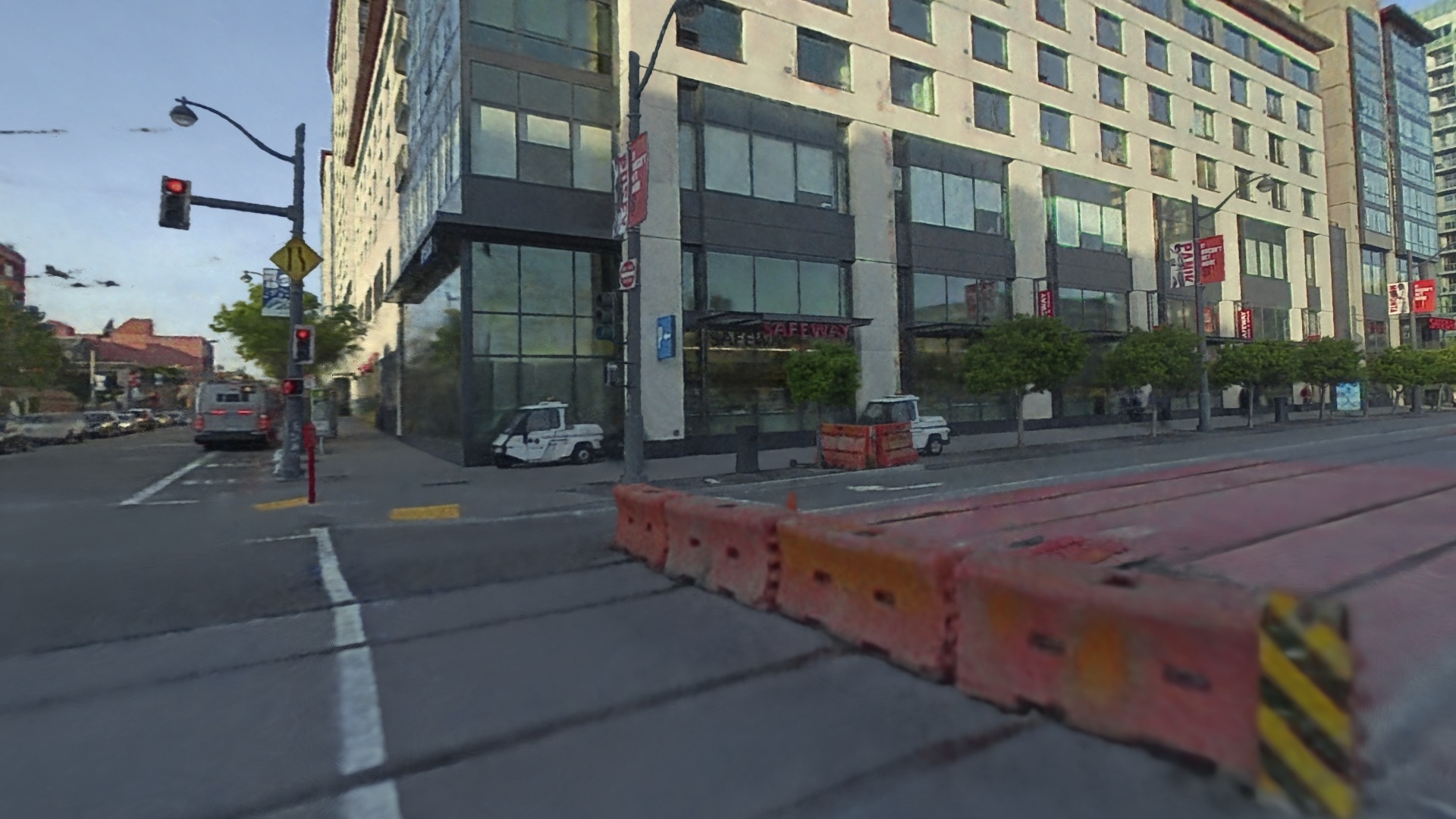}};
            \begin{scope}[shift={(img.south west)},x={(img.south east)},y={(img.north west)}]
                \spy [red,magnification=2.5,size=1.3cm] on (-1.7,0.75) in node at (-1.2,-0.5);
                \spy [orange,magnification=4,size=1cm] on (1.35,0.45) in node at (0.99,-0.5);
            \end{scope}
        \end{tikzpicture}%
        }
        \caption{MOISST++ (Nerfacto)}
        \label{figures/pandaset/qualitative_ns/moisst_nerfacto}
    \end{subfigure}
    \hfill
    \begin{subfigure}{.48\linewidth}
        \centering
        \resizebox{\linewidth}{!}{%
        \begin{tikzpicture}[outer sep=0pt,inner sep=0pt,spy using outlines={rectangle, connect spies}]
            \node (img) {\includegraphics[width=\linewidth]{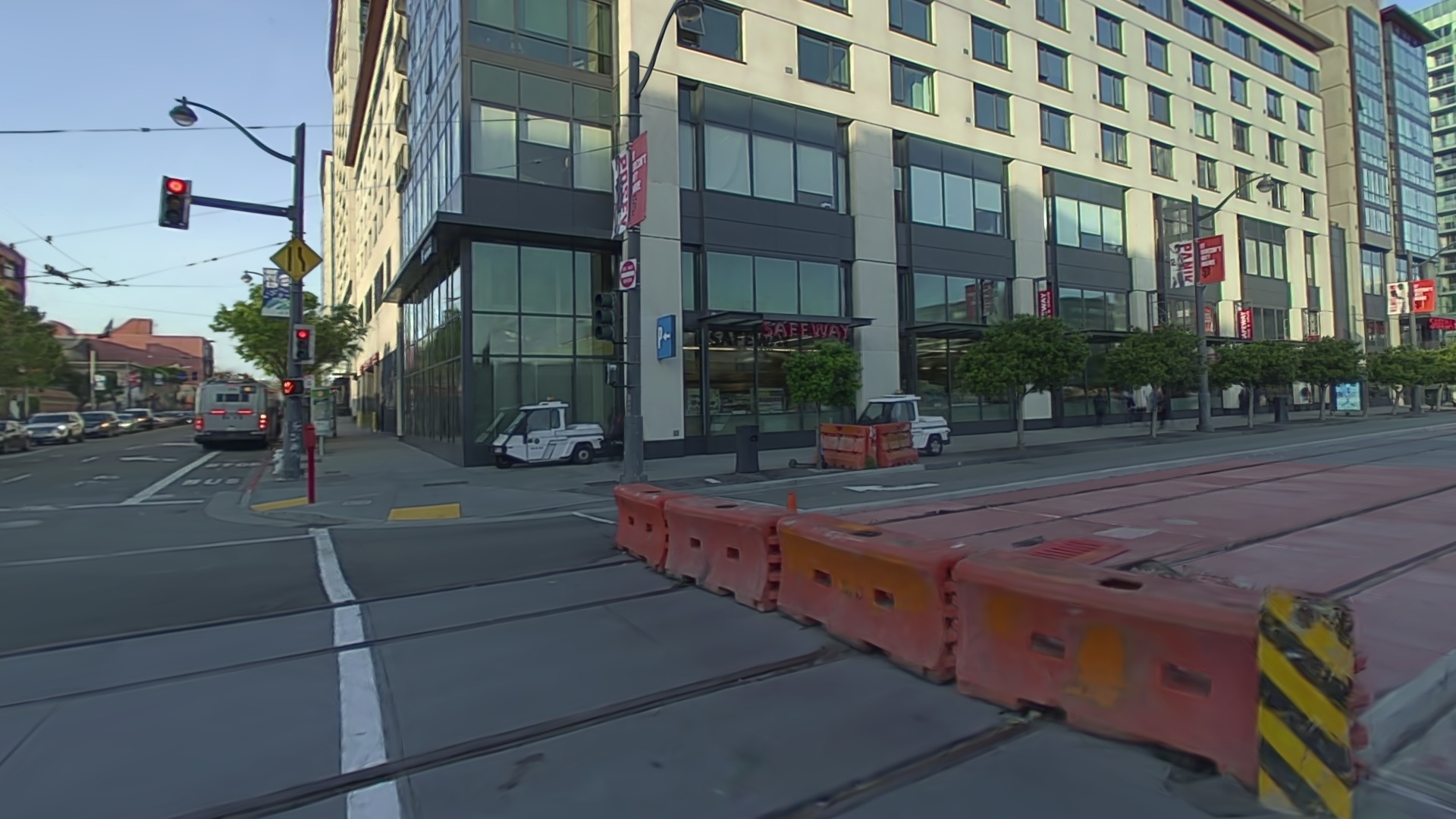}};
            \begin{scope}[shift={(img.south west)},x={(img.south east)},y={(img.north west)}]
                \spy [red,magnification=2.5,size=1.3cm] on (-1.7,0.75) in node at (-1.2,-0.5);
                \spy [orange,magnification=4,size=1cm] on (1.35,0.45) in node at (0.99,-0.5);
            \end{scope}
        \end{tikzpicture}%
        }
        \caption{MOISST++ (Splatfacto)}
        \label{figures/pandaset/qualitative_ns/moisst_splatfacto}
    \end{subfigure}

    \vspace{0.3mm}

    \caption{\textbf{PandaSet qualitative (Nerfstudio):} GT image and renderings with original vs refined poses (MOISST++), shown with both Nerfacto and Splatfacto.}
    \label{figures/pandaset/qualitative_ns}
\end{figure}

\begin{figure}[!t]
    \centering
    \setlength{\tabcolsep}{0.0001\linewidth}
    \begin{subfigure}{\linewidth}
        \centering
        \begin{tabular}{ccc}
            \includegraphics[width=.333\linewidth]{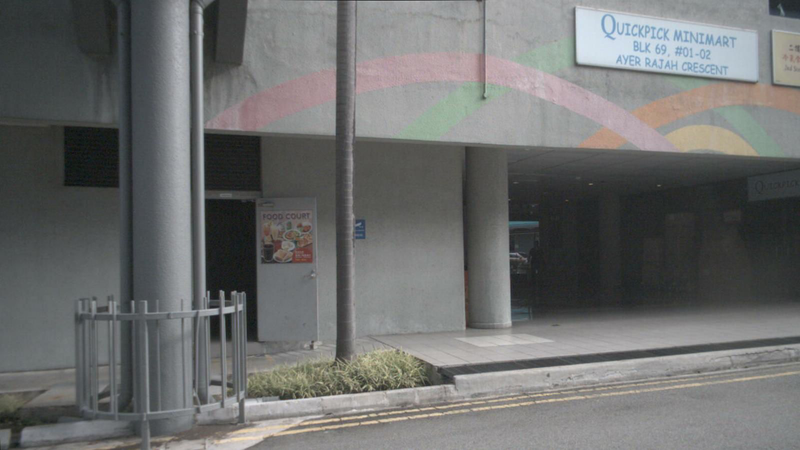} &
            \includegraphics[width=.333\linewidth]{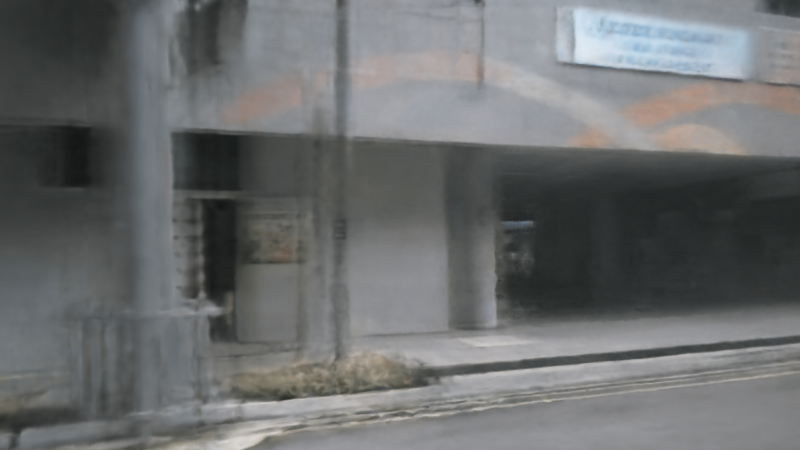} &
            \includegraphics[width=.333\linewidth]{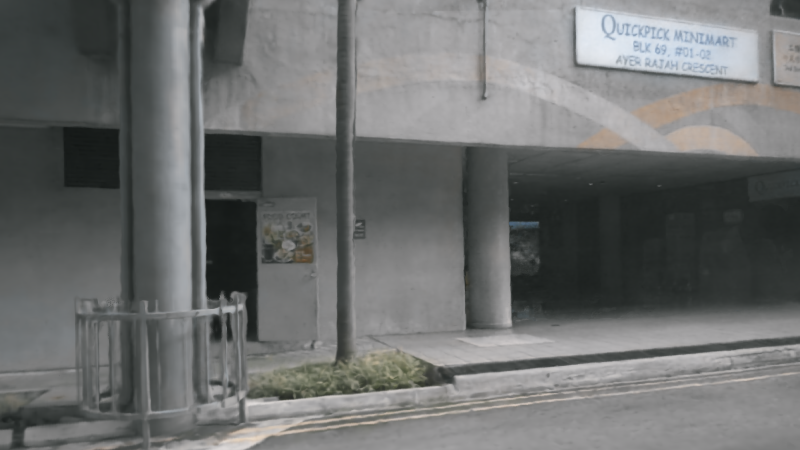}
        \end{tabular}
        \vspace{-1mm}
        \caption{RGB renderings (GT / Original / MOISST++)}
        \label{figures/nuscenes/rgb_normals/rgb}
    \end{subfigure}

    \vspace{0.3mm}

    \begin{subfigure}{\linewidth}
        \centering
        \begin{tabular}{ccc}
            \phantom{\includegraphics[width=.333\linewidth]{ressources/nuScenes/qualitative/3_1174_gt.png}} &
            \includegraphics[width=.333\linewidth]{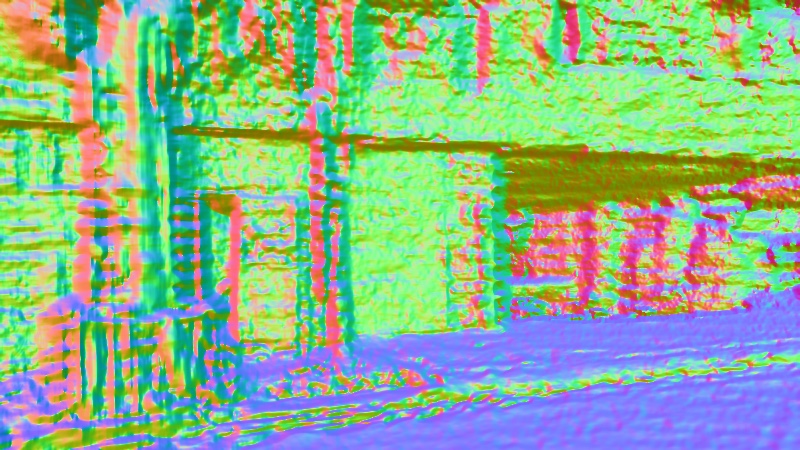} &
            \includegraphics[width=.333\linewidth]{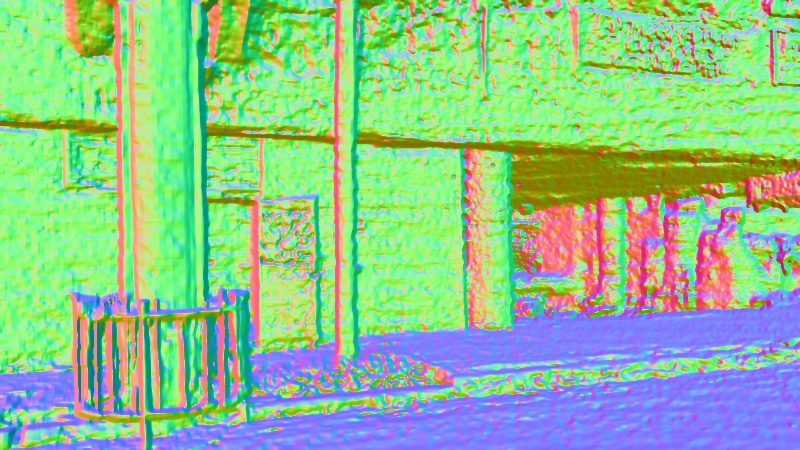}
        \end{tabular}
        \vspace{-1mm}
        \caption{Normal maps (Original / MOISST++)}
        \label{figures/nuscenes/rgb_normals/normals}
    \end{subfigure}

    \caption{\textbf{nuScenes qualitative (NeRF-LS):} original vs refined poses. MOISST++ reduces pose-induced artifacts and yields sharper reconstructions.}
    \label{figures/nuscenes/qualitative_rgb_normals}
\end{figure}

For NVS, NeRF-LS is trained with 1 out of 2 frames and evaluated on the held-out remainder; Nerfstudio uses its default interval split (1 out of 8 for evaluation).
PSNR, SSIM, and LPIPS are reported in Tables~\ref{tab:NVS_imgine},~\ref{tab:NVS_Nerfacto} and~\ref{tab:NVS_splatfacto}. MOISST++ achieves the best results on all NVS metrics and datasets. The improvement over the original poses is largest where the original calibration is weakest, e.g.\ on nuScenes the NeRF-LS PSNR rises from $22.34$ to $23.46$ and LPIPS drops from $0.465$ to $0.339$, while on the already-accurate KITTI-360 it is marginal ($22.42\!\to\!22.61$ PSNR). The neural-rendering competitor SOAC++ is most often the second-best method, ahead of the camera-only Nerfstudio baseline; both generally improve the NVS metrics over the original poses, but neither matches the unified LiDAR-and-camera representation of MOISST++. The gains hold across three renderers spanning NeRF backends (NeRF-LS, Nerfacto) and 3D Gaussian Splatting (Splatfacto; e.g.\ nuScenes PSNR $22.29\!\to\!25.66$), showing they are not tied to a specific reconstruction method.
Qualitative results are shown for nuScenes (Fig.~\ref{figures/nuscenes/qualitative_rgb_normals}) and PandaSet (Fig.~\ref{figures/pandaset/qualitative_ns}).

\subsection{Geometric consistency}
\begin{figure}[!t]
    \centering
    \setlength{\tabcolsep}{0.0001\linewidth}
    \begin{tabular}{ccc}
        \includegraphics[width=.333\linewidth,trim={0 3cm 0 3cm},clip=true]{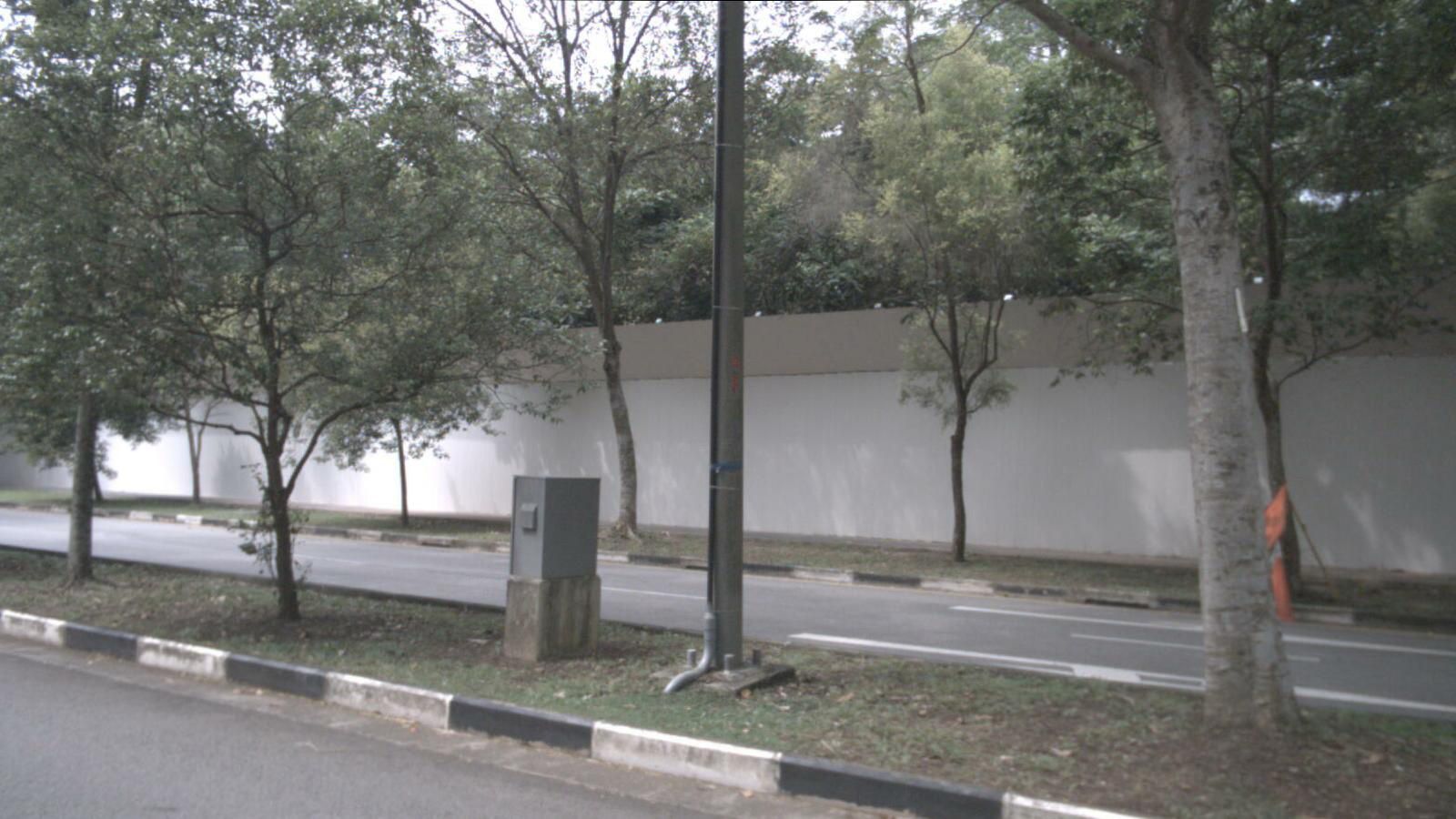} &
        \includegraphics[width=.333\linewidth,trim={0 4.5cm 0 4.5cm},clip=true]{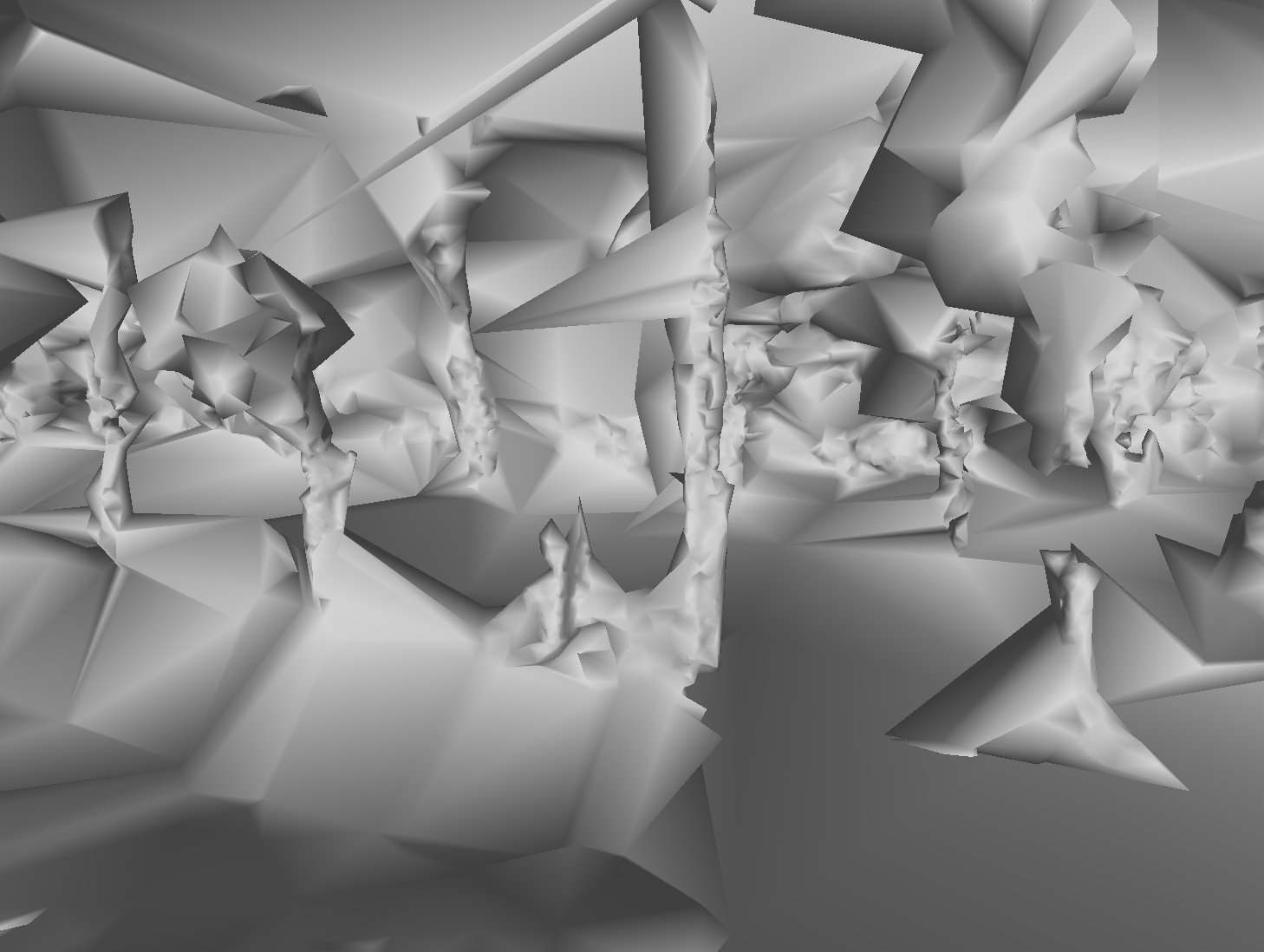} &
        \includegraphics[width=.333\linewidth,trim={0 4.5cm 0 4.5cm},clip=true]{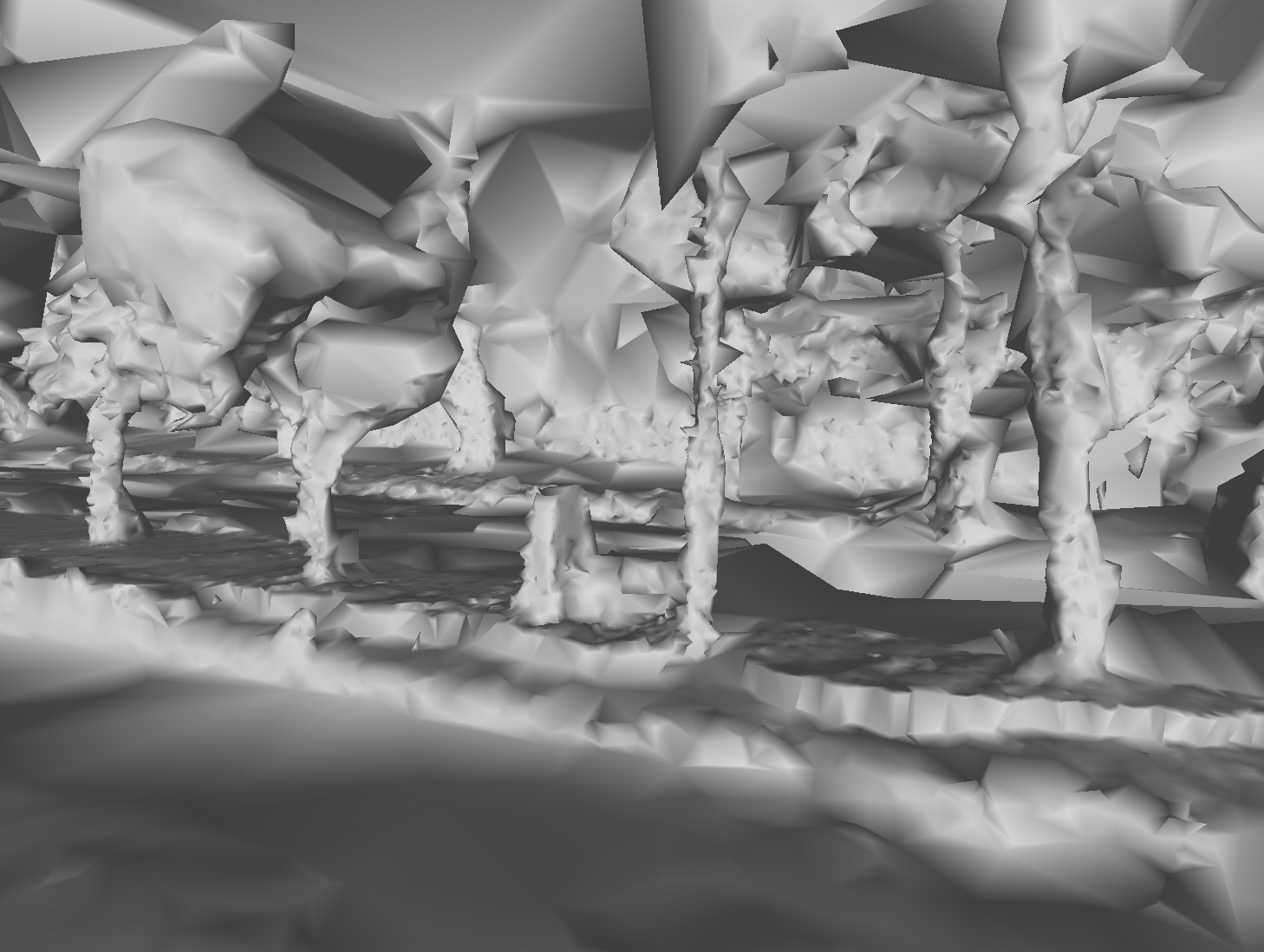}\\
        {\footnotesize (a) GT} & {\footnotesize (b) Original} & {\footnotesize (c) MOISST++}
    \end{tabular}
    \caption{\textbf{nuScenes qualitative (Delaunay mesh):} meshes reconstructed from images with original vs refined poses; refined poses yield cleaner road surfaces and sharper thin structures.}
    \label{figures/nuscenes/delauney_compact}
\end{figure}

Table~\ref{tab:delauney_metrics} reports surface metrics between the Delaunay mesh and the accumulated LiDAR point cloud. This check is not circular: although LiDAR is used during optimization, the mesh is reconstructed from images only (COLMAP MVS and Delaunay meshing), so the metric measures how well this image-derived surface agrees with the independent LiDAR geometry, i.e.\ how well camera and LiDAR poses are aligned. MOISST++ improves camera-LiDAR alignment across all datasets, with the largest gains on nuScenes and Waymo where original calibration errors are more pronounced. Fig.~\ref{figures/nuscenes/delauney_compact} illustrates the mesh quality difference on nuScenes: refined poses produce smoother road surfaces and sharper thin structures such as tree trunks and poles, whereas the original poses lead to noisy, fragmented geometry.

\begin{table*}[t]
    \centering
        \caption{Additional experiments on PandaSet (mean over 10 subsequences), reported on the same GT-free metrics as Table~\ref{tab:metrics}. Prec.\ is the Delaunay precision at $15$\,cm; P$\rightarrow$M is the point-to-mesh distance; NVS metrics are NeRF-LS.}
    \label{tab:experiments}

    \scriptsize
    \setlength{\tabcolsep}{4pt}
    \renewcommand{\arraystretch}{1.0}

    \begin{subtable}[t]{\textwidth}
        \centering
        \caption{Controlled perturbation on PandaSet (frozen, no optimization). Known errors of increasing magnitude are injected into the Original poses; every metric degrades monotonically and per-frame trajectory error dominates. The first row is the unperturbed Original (Table~\ref{tab:metrics}).}
        \label{tab:perturb}

        \begin{tabular}{l @{\hspace{8pt}} ccccccc}
        \toprule
        Injected error & Reproj.~$\downarrow$ & Track~$\uparrow$ & PSNR~$\uparrow$ & SSIM~$\uparrow$ & LPIPS~$\downarrow$ & Prec.~$\uparrow$ & P$\rightarrow$M~[m]~$\downarrow$\\
        \midrule
        None (Original poses)                     & \textbf{1.19}  & \textbf{6.32} & \textbf{22.08} & \textbf{0.788} & \textbf{0.482} & \textbf{0.385} & \textbf{0.378}\\
        Calibration $1^\circ$/$0.10$\,m           & 1.294 & 5.88 & 21.60 & 0.771 & 0.582 & 0.23 & 0.69\\
        Calibration $2^\circ$/$0.20$\,m           & 1.432 & 5.68 & 21.25 & 0.762 & 0.630 & 0.13 & 1.18\\
        Trajectory $\pm5^\circ$/$\pm0.15$\,m      & 1.810 & 2.01 & 17.75 & 0.719 & 0.857 & 0.00 & 7.50\\
        Calibration $2^\circ$/$0.20$\,m $+$ traj. & 1.818 & 2.01 & 17.72 & 0.719 & 0.858 & 0.10 & 7.53\\
        \bottomrule
        \end{tabular}
        
    \end{subtable}

    \vspace{7pt}

    \begin{subtable}[t]{\textwidth}
        \centering
        \caption{Optimization ablations on PandaSet. Top: single optimization stages vs.\ the full pipeline (third row, the default configuration). Bottom row: full pipeline with extrinsics estimated per subsequence instead of shared; the accuracy is near-identical, so a single shared calibration is used by default.}
        \label{tab:ablation}

        \begin{tabular}{l @{\hspace{8pt}} ccccccc}
        \toprule
        Variant & Reproj.~$\downarrow$ & Track~$\uparrow$ & PSNR~$\uparrow$ & SSIM~$\uparrow$ & LPIPS~$\downarrow$ & Prec.~$\uparrow$ & P$\rightarrow$M~[m]~$\downarrow$\\
        \midrule
        Calibration only                          & 1.001 & 6.87 & 22.08 & 0.790 & 0.489 & 0.36 & 0.36\\
        Trajectory only                           & 1.109 & 6.50 & 22.04 & 0.790 & 0.501 & 0.41 & 0.36\\
        Full (calib.\,$+$\,traj., shared extr.)   & 1.004 & 6.86 & 22.15 & 0.796 & 0.471 & 0.38 & 0.36\\
        Full, per-subsequence extr.               & 0.992 & 6.86 & 22.25 & 0.799 & 0.465 & 0.43 & 0.32\\
        \bottomrule
        \end{tabular}
        
    \end{subtable}

\end{table*}

\begin{table*}[t]
    \centering
        \caption{Per-subsequence statistical significance, MOISST++ vs.\ the Original dataset poses. Each cell reports the mean\,$\pm$\,std of the per-subsequence difference $\Delta$ (MOISST++ $-$ Original), the win/tie/loss counts (number of subsequences that improve/tie/degrade, summing to $n$), and the paired Wilcoxon signed-rank $p$-value. Entries marked $^{\ast}$ are not significant at the $0.05$ level; all other reported results are significant.}
    \label{tab:significance}

    \scriptsize
    \setlength{\tabcolsep}{4pt}
    \renewcommand{\arraystretch}{1.0}
    \resizebox{\textwidth}{!}{%
    \begin{tabular}{l cccc}
    \toprule
    Metric & KITTI-360 ($n{=}50$) & nuScenes ($n{=}10$) & PandaSet ($n{=}10$) & Waymo ($n{=}10$)\\
    \midrule
    COLMAP reproj.\ (px)~$\downarrow$ & $-0.010{\pm}0.055$,\; $24/0/26$,\; $p{=}0.79^{\ast}$ & $-0.488{\pm}0.210$,\; $10/0/0$,\; $p{=}0.006$ & $-0.206{\pm}0.049$,\; $10/0/0$,\; $p{=}0.006$ & $-0.122{\pm}0.069$,\; $10/0/0$,\; $p{=}0.006$\\
    COLMAP track~$\uparrow$          & $+0.155{\pm}0.624$,\; $41/0/9$,\; $p{<}10^{-4}$   & $+1.796{\pm}1.640$,\; $10/0/0$,\; $p{=}0.006$ & $+0.537{\pm}0.288$,\; $10/0/0$,\; $p{=}0.006$ & $+0.384{\pm}0.388$,\; $10/0/0$,\; $p{=}0.006$\\
    NeRF-LS PSNR~$\uparrow$          & $+0.196{\pm}0.269$,\; $43/0/7$,\; $p{<}10^{-4}$   & $+1.125{\pm}0.593$,\; $10/0/0$,\; $p{=}0.006$ & $+0.278{\pm}0.134$,\; $10/0/0$,\; $p{=}0.006$ & $+0.476{\pm}0.251$,\; $10/0/0$,\; $p{=}0.006$\\
    Nerfacto PSNR~$\uparrow$         & $+0.126{\pm}0.310$,\; $38/1/11$,\; $p{=}0.002$     & $+1.270{\pm}0.482$,\; $10/0/0$,\; $p{=}0.006$ & $+0.807{\pm}0.143$,\; $10/0/0$,\; $p{=}0.006$ & $+1.002{\pm}0.472$,\; $10/0/0$,\; $p{=}0.006$\\
    Splatfacto PSNR~$\uparrow$       & $+0.229{\pm}0.704$,\; $34/1/15$,\; $p{=}0.004$     & $+3.373{\pm}1.323$,\; $10/0/0$,\; $p{=}0.006$ & $+1.093{\pm}0.263$,\; $10/0/0$,\; $p{=}0.006$ & $+1.793{\pm}0.699$,\; $10/0/0$,\; $p{=}0.006$\\
    Delaunay prec.~$\uparrow$        & $+0.011{\pm}0.012$,\; $33/14/3$,\; $p{<}10^{-4}$   & $+0.251{\pm}0.116$,\; $10/0/0$,\; $p{=}0.006$ & $+0.015{\pm}0.018$,\; $7/2/1$,\; $p{=}0.052^{\ast}$ & $+0.185{\pm}0.112$,\; $10/0/0$,\; $p{=}0.006$\\
    Delaunay P$\rightarrow$M (m)~$\downarrow$ & $-0.026{\pm}0.055$,\; $38/10/2$,\; $p{<}10^{-4}$ & $-0.167{\pm}0.301$,\; $10/0/0$,\; $p{=}0.006$ & $-0.023{\pm}0.081$,\; $7/0/3$,\; $p{=}0.61^{\ast}$ & $-0.068{\pm}0.046$,\; $9/1/0$,\; $p{=}0.009$\\
    \bottomrule
    \end{tabular}}
    
\end{table*}

\subsection{Controlled perturbation}
Because the evaluation is ground-truth-free, we verify that the metrics react correctly to \emph{known} pose error. Starting from the Original PandaSet poses, we inject controlled $SE(3)$ errors and pass the perturbed poses straight to the metrics \emph{without} any optimization: a constant per-camera extrinsic offset (miscalibration, $1^\circ$/$0.10$\,m, then $2^\circ$/$0.20$\,m), a per-frame trajectory perturbation ($\pm5^\circ$/$\pm0.15$\,m), and their combination. As shown in Table~\ref{tab:perturb}, every metric family degrades monotonically with the injected error, and trajectory error dominates: relative to the Original poses the reprojection error rises by about $50\%$, the track length collapses from $6.3$ to $2.0$, PSNR drops by more than $4$\,dB, and the point-to-mesh distance grows from $0.38$\,m to several metres. This confirms the metrics are sensitive to genuine calibration and trajectory error.

\subsection{Ablation studies}
Table~\ref{tab:ablation} ablates the two optimization stages on PandaSet. Calibration alone recovers most of the triangulation gain (reprojection $1.19\!\to\!1.00$), while the trajectory correction $\mathcal{T}_{cor}$ alone mainly improves geometric consistency (precision $0.385\!\to\!0.41$); only their combination also improves the NVS metrics (LPIPS $0.489\!\to\!0.471$ over calibration-only) and the track length, confirming the two stages are complementary. Table~\ref{tab:ablation} also compares a single shared extrinsic calibration against estimating extrinsics independently per subsequence. Per-subsequence estimation is marginally ahead on every metric, but the gaps ($0.10$\,dB PSNR, $0.01$\,px reprojection) are negligible against the cross-sequence spread. A shared calibration therefore costs essentially nothing in accuracy while being more robust, as pooling geometric evidence across subsequences remains reliable where per-subsequence estimation would be under-constrained; we adopt it as the default whenever the rig is physically fixed. On Waymo, where the data aggregates different capture vehicles, extrinsics are instead optimized per subsequence and the shared-calibration stage is inactive (Sec.~\ref{setup/subseq}).

\subsection{Discussion}
MOISST++ remains best across metrics. SOAC++ improves over the original poses on nearly every metric and is generally second-best, while the unconstrained camera-only Nerfstudio optimization is not always beneficial, even degrading the reprojection error below the original poses on KITTI-360 ($0.586{\to}0.691$) and Waymo ($1.137{\to}1.466$). For already-reasonable dataset poses, the rig-constrained unified representation of MOISST++ converges more reliably than either the partitioned (SOAC++) or unconstrained (Nerfstudio) alternative.

KITTI-360 shows the smallest improvements, reflecting its already high-quality poses obtained through careful SLAM and loop closure optimization. nuScenes benefits the most due to its missing Z-axis information (see Table~\ref{tab:metrics}). PandaSet and Waymo fall in between, with improvements across most metric families, suggesting that their original localization pipelines leave room for refinement especially on extrinsic calibration.

\textbf{Statistical significance.} We assess significance per subsequence (KITTI-360 $n{=}50$; nuScenes, PandaSet, Waymo $n{=}10$) with a paired Wilcoxon signed-rank test between MOISST++ and the Original poses, reported with the mean\,$\pm$\,std of the per-subsequence differences and win/tie/loss counts in Table~\ref{tab:significance}. On nuScenes and Waymo every metric family is significant in (nearly) every subsequence, and on PandaSet the triangulation and NVS metrics are significant throughout. On KITTI-360 the gains are smaller but the track length, NVS and geometric-consistency metrics remain significant, whereas the reprojection error is statistically indistinguishable from the Original ($24/0/26$, $p{=}0.79$), consistent with its already-accurate poses. The only remaining non-significant case is the Delaunay geometric-consistency on PandaSet (precision $p{=}0.052$; P$\rightarrow$M $p{=}0.61$): MOISST++ improves it on average and does not degrade it. The per-subsequence spread of the raw metrics is similar for \emph{all} methods and dominated by subsequence difficulty rather than by the method: e.g.\ the Nerfacto PSNR std is $2.1/2.0/2.2/2.0$\,dB for Original/Nerfstudio/SOAC++/MOISST++ on KITTI-360 and $0.87/0.79/0.93/0.87$\,dB on PandaSet. Since this shared spread exceeds the mean gaps of Table~\ref{tab:metrics}, aggregate std is uninformative and we rely on the paired statistics above; notably, MOISST++ also shows the smallest reprojection-error spread on every dataset (e.g.\ $0.019$ vs.\ $0.029$--$0.061$\,px std on PandaSet), i.e.\ the most consistent poses.

\textbf{Cross-metric consistency.} To confirm the metric families tell a coherent story, we measure for every metric pair and subsequence whether both agree on the direction of change induced by MOISST++. Averaged over all pairs, sign agreement is $100\%$ on nuScenes, $98\%$ on Waymo, $91\%$ on PandaSet, and $68\%$ on KITTI-360. The lower, mixed agreement on KITTI-360 is expected given its near-noise-level changes; the high agreement on the other three datasets ($91$--$100\%$), across heterogeneous metrics (triangulation, NVS, geometric consistency), indicates the gains reflect genuine pose-quality improvements rather than metric-specific artifacts.

\subsection{Limitations}
The benefit of the refinement is not uniform across subsequences. On the already-accurate KITTI-360, the reprojection error improves on only about half of the subsequences ($24/0/26$) and the change is not significant, and the PandaSet geometric-consistency metrics show no significant per-subsequence change. In these cases MOISST++ leaves the poses essentially unchanged rather than degrading them, but cannot add information where the original annotations are already near the accuracy ceiling.

Since pose optimization relies on NeRF-based reconstruction, it inherits NeRF's limitations: degraded geometry in night scenes and in environments with repetitive or absent features (tunnels, long walls), where LiDAR supervision only partially mitigates divergence; far-field structures beyond the reconstructed volume are also poorly modeled. Future work could explore reconstruction backends better suited to these degenerate conditions, handle temporally varying calibration for rigs subject to mechanical drift, validate the refined poses against survey-grade references on multimodal rigs outside the driving domain, and benchmark against strong non-NeRF LiDAR-inertial odometry and pose-graph baselines.

\section{Conclusion}
This work introduces MOISST++, a pipeline for optimizing sensor poses and calibration parameters in autonomous driving datasets, accompanied by a rigorous multi-metric evaluation framework. Our experiments show statistically significant improvements on nuScenes, PandaSet and Waymo and marginal changes on the already well-calibrated KITTI-360, validated by triangulation, novel view synthesis, and geometric consistency metrics that show strong cross-metric agreement, and a controlled-perturbation study confirming the metrics respond to known pose error. These refined poses directly benefit downstream tasks such as sensor fusion, 3D reconstruction, and localization.

The results also highlight that evaluation data itself may contain inherent inaccuracies: when metric variations are minimal (as for KITTI-360), this may reflect the high precision of the underlying dataset rather than a lack of improvement, underscoring the importance of refining both methods and dataset quality for reliable benchmarking.

By publicly releasing the optimized poses and calibration parameters, we aim to foster more robust models and more reliable evaluations across the field.

\FloatBarrier
\bibliographystyle{IEEEtran}
\bibliography{IEEEabrv,cvpr_biblio}

\end{document}